\pdfoutput=1

\documentclass[11pt]{article}

\usepackage{ACL2023}

\usepackage{times}
\usepackage{latexsym}

\usepackage[T1]{fontenc}

\usepackage[utf8]{inputenc}

\usepackage{microtype}

\usepackage{inconsolata}
\usepackage{enumitem}
\setlist[enumerate]{topsep=0pt,itemsep=-1ex,partopsep=1ex,parsep=1ex}
\newcommand*\samethanks[1][\value{footnote}]{\footnotemark[#1]}
\usepackage{pgfplots} 
\pgfplotsset{compat=1.13}
\newcommand{\minisection}[1]{\noindent{\bf #1}\hspace{0.6em}}

\usepackage{graphicx} 
\usepackage{url}
\usepackage{booktabs}
\usepackage{float}
\usepackage{rotating}
\definecolor{lightred}{HTML}{CC8685}
\definecolor{darkred}{HTML}{FF0000}

\title{Exploring the Curious Case of Code Prompts}

\author{Li Zhang\thanks{~~Equal contribution.}, \quad
  Liam Dugan\samethanks, \quad
  Hainiu Xu\samethanks, \quad
  Chris Callison-Burch \\
  University of Pennsylvania \\
  {\tt \{zharry,ldugan,seacow,ccb\}@seas.upenn.edu}
}

\begin{document}
\maketitle
\begin{abstract}
Recent work has shown that prompting language models with code-like representations of natural language leads to performance improvements on structured reasoning tasks. However, such tasks comprise only a small subset of all natural language tasks. In our work, we seek to answer whether or not code-prompting is the preferred way of interacting with language models \textit{in general}. We compare code and text prompts across three popular GPT models (\texttt{davinci}, \texttt{code-davinci-002}, and \texttt{text-davinci-002}) on a broader selection of tasks (e.g., QA, sentiment, summarization) and find that with few exceptions, code prompts do not consistently outperform text prompts. Furthermore, we show that the style of code prompt has a large effect on performance for some but not all tasks and that fine-tuning on text instructions leads to better relative performance of code prompts.
\end{abstract}

\begin{table*}[ht]
    \small
    \centering
    \begin{tabular}{l|l|l|l|l}
    \toprule
    Dataset & Task Category & Num. Eval Examples & Metric & Origin \\
    \midrule
    HellaSwag &Commonsense Reasoning&1000 / 10042&Accuracy&\citet{zellers2019hellaswag}  \\
    wikiHow Goal-Step &Commonsense Reasoning&1000 / 1073&Accuracy&\citet{zhang2020reasoning}  \\
    wikiHow Temporal &Commonsense Reasoning&1000 / 3100&Accuracy&\citet{zhang2020reasoning} \\
    WinoGrande &Commonsense Reasoning&1000 / 1767&Accuracy&\citet{sakaguchi2021winogrande} \\
    OpenPI &Commonsense Reasoning&111 / 111&ROUGE-F1&\citet{tandon2020dataset}\\
    ANLI & Natural Language Inference &1000 / 3000&Accuracy&\citet{nie2020adversarial} \\
    Yelp &Sentiment Analysis&1000 / 10000&Pearson's r&\citet{zhang2015character} \\
    IMDb &Sentiment Analysis&1000 / 25000 &Accuracy&\citet{maas2011learning}\\
    HotpotQA &Question Answering&1000 / 7405&Macro-F1 &\citet{yang2018hotpotqa} \\
    SQuAD &Question Answering&1000 / 11873&Macro-F1 &\citet{rajpurkar2018know} \\
    CNN/Daily Mail &Summarization&1000 / 13368&ROUGE-2&\citet{nallapati2016abstractive}\\
    XSUM &Summarization&1000 / 11332&ROUGE-2&\citet{narayan2018don}\\
    \bottomrule
    \end{tabular}
    \caption{The 12 evaluation tasks. Macro F1 is based on \citet{rajpurkar2016squad}. For each task, we randomly sample a fixed set of 1000 examples from its validation or test set for evaluation. For OpenPI we are limited to 111 examples.}
    \label{tab:all-tasks}
\end{table*}

\section{Introduction}
Recent work has shown that pre-training language models (LMs) on a mixture of text and program code (e.g., Python or Javascript) makes them more capable of reasoning over natural language \cite{suzgun2022challenging}.
Such program-trained language models (PLMs) significantly outperform text-only LMs on tasks such as math problems and tracking shuffled objects despite such tasks lacking any explicit code formulae \cite{liang2022holistic}.

\begin{figure}
    \centering
    \includegraphics[width=\columnwidth]{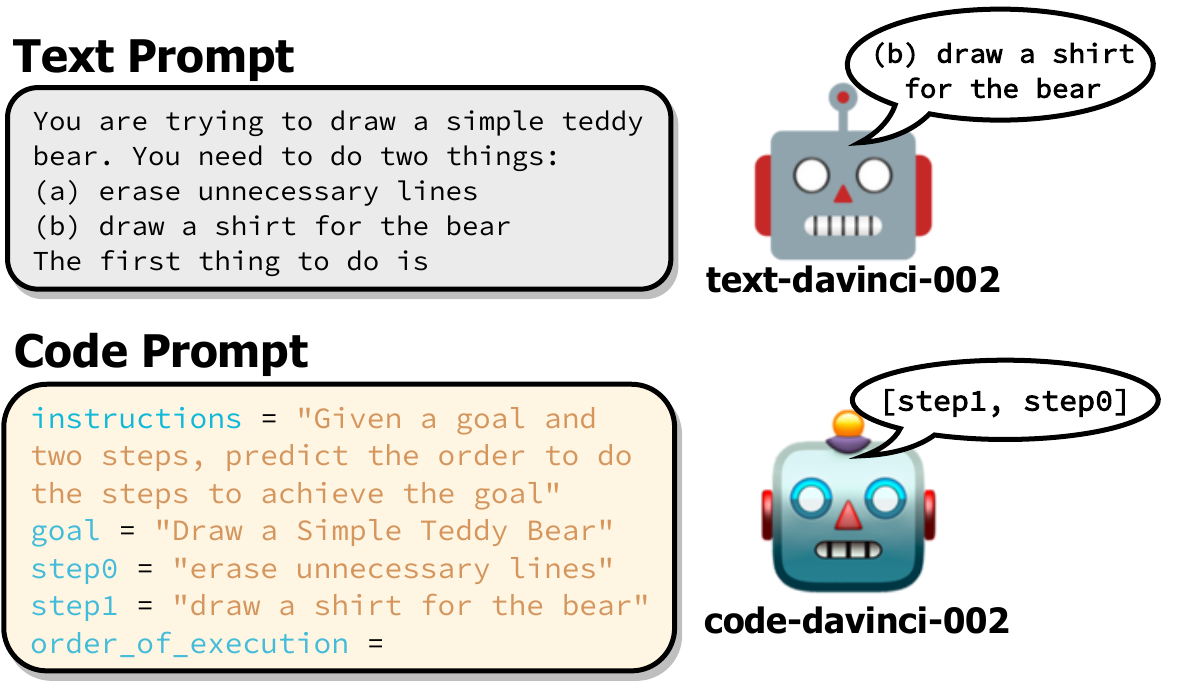}\hspace*{-0.5cm}
    \caption{For certain tasks, prompting program-trained language models with code\textit{-like} representations works better than prompting with text.}
    \label{fig:page1}
\end{figure}

Furthermore, prompting such PLMs with \textit{code-like structures} (e.g., Python, JSON, PDDL) instead of text has been shown to lead to performance improvements on structured common sense reasoning \cite{madaan2022language}, event argument extraction \cite{wang2022code4struct}, knowledge graph construction \cite{bi2023codekgc}, story understanding \cite{dong2022corrpus}, and causal reasoning \cite{zhang-etal-2023-causal}.

Such results naturally lead us to ask whether code-prompting is the preferred way of interacting with PLMs \textit{in general}. While previous work is limited to reasoning tasks, in this work we analyze a broad selection of tasks (e.g., QA, sentiment, summarization) and systematically compare the performance of prompting PLMs with code vs. prompting with text\footnote{The code, prompts, and outputs for our experiments are public at \url{https://github.com/zharry29/codex_vs_gpt3}}.
We find that:
\begin{itemize}[noitemsep,nolistsep]
    \item With the exception of some reasoning tasks, code prompts do not outperform text prompts
    \item The style of code prompt has a large effect on performance for some but not all tasks.
    \item Fine-tuning on text instructions leads to relative improvements when using code prompts.
\end{itemize}

\section{Experimental Design}
\label{sec:experimental_design}
\paragraph{Model Selection}
For our text-based LM we use the original 175 billion parameter \texttt{davinci} model introduced by \citet{brown2020fewshot}. For our PLM we use the newer \texttt{code-davinci-002} model which was explicitly trained on text and code. Neither model underwent any supervised instruction fine-tuning. In addition, we analyze performance on \texttt{text-davinci-002}, which is a variant of \texttt{code-davinci-002} trained explicitly on human demonstrations using supervised fine-tuning\footnote{\url{https://platform.openai.com/docs/model-index-for-researchers}}. We include this model to help us determine whether or not fine-tuning PLMs on text instructions affects their ability to interpret code prompts. All three models were queried through the OpenAI API\footnote{\url{https://openai.com/blog/openai-api}} and our experiments cost approximately \$2700 in total (see Appendix \ref{app:eval_cost} for the full cost breakdown).

\paragraph{Task Selection}
Following the methodology of \citet{sanhmultitask} we select tasks in a top-down fashion by first choosing the categories of interest (e.g. Question Answering, Sentiment Analysis, Summarization) and then selecting datasets from within those categories. We pay special attention to common sense and causal reasoning tasks as PLMs prompted with code have been shown to perform well on such tasks. The resulting 12 tasks are listed in Table~\ref{tab:all-tasks} and include Commonsense Reasoning, Natural Language Inference, Sentiment Analysis, Question Answering, and Summarization. More details on each task can be found in Appendix \ref{sec:detailed-tasks}.

\paragraph{Prompt Formulation}
We collect text prompts for each task using the PromptSource dataset \citep{bach2022promptsource}, a publicly available collection of crowd-sourced prompt templates. For tasks with many prompts, we manually select one from those provided in the dataset. For a few tasks absent on PromptSource, we write the prompts ourselves.

For our code prompts, we manually write four custom code prompts per task. The code prompt types are as follows, from least to most Pythonic.

\begin{figure*}
    \centering
    \includegraphics[width=2\columnwidth]{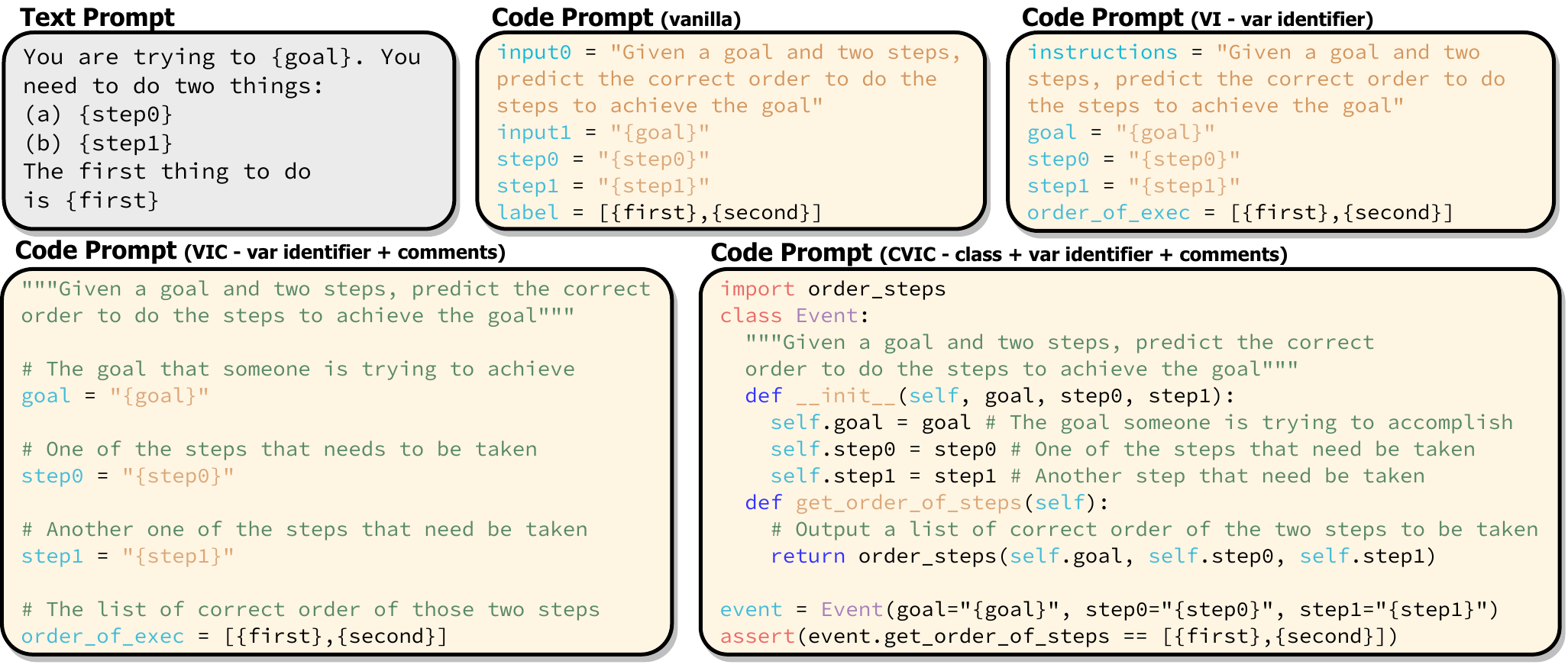}
    \caption{An example of the four styles of manually written code prompts used in our analysis (\texttt{Vanilla}, \texttt{VI}, \texttt{VIC}, and \texttt{CVIC}) for the wikiHow temporal ordering task. At test time, variables in braces are replaced with information from the dataset item (as shown in Figure \ref{fig:page1}). For this task, \texttt{\{goal\}}, \texttt{\{step0\}}, \texttt{\{step1\}} refer to the article title and the steps to order while \texttt{\{first\}} and \texttt{\{second\}} refer to the true ordering of the steps.}
    \label{fig:code-prompts}
\end{figure*}

\renewcommand{\theenumi}{\textbf{(\roman{enumi})}}
\begin{enumerate}[noitemsep,nolistsep]
    \item \textbf{Vanilla (\texttt{Vanilla})}: instructions and inputs are given as variables with generic names;
    \item \textbf{Var Identifier (\texttt{VI})}: instructions and inputs are given as variables with meaningful names;
    \item \textbf{Var Identifier + Comments (\texttt{VIC})}: instructions and inputs are given as variables with meaningful names along with comments explaining their purpose;
    \item \textbf{Class + Var Identifier + Comments (\texttt{CVIC})}: instructions and inputs are given as a task-specific \texttt{class}. Functionality is ``implemented'' as member functions.
\end{enumerate}

Figure \ref{fig:code-prompts} shows an example of the different styles of code prompts for the wikiHow temporal ordering task. Note that we attempt to write our code prompts such that we match the wording of the text-based PromptSource prompt as closely as possible.

At inference time, for each test example, we randomly sample in-context examples from the training set and add them to the context window until the maximum context length is reached. This process circumvents the bias caused by static in-context examples. We conduct an ablation study where we vary the random seed and show that this process produces consistent results (see Appendix~\ref{app:ablation_study}).

\section{Results}

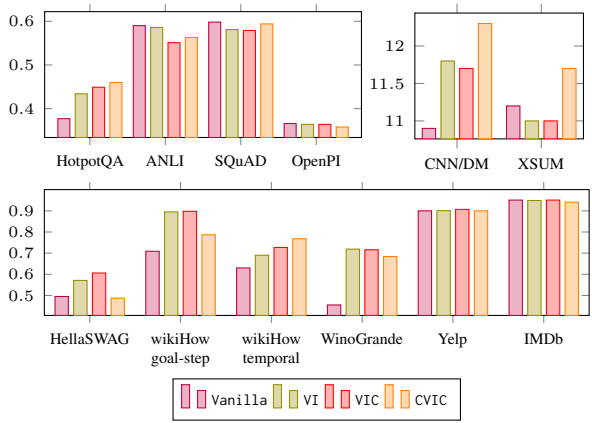
\begin{figure}[t!]
\centering
\begin{tikzpicture}
    \begin{axis}[ybar,width=5.75cm,height=3.25cm,enlarge x limits=0.2,
        symbolic x coords={HotpotQA,ANLI,SQuAD,OpenPI},xtick=data,yticklabel style={overlay,font=\tiny},legend style={font=\tiny},bar width=4.5pt,xticklabel style={text width=1.5cm,font=\tiny,align=center,max space between ticks=12,major tick length=0.1cm,},]
        \addplot[purple,fill=purple!30!white]
        coordinates{(HotpotQA,0.377) (ANLI,0.590) (SQuAD,0.598) (OpenPI,0.366)};
        \addplot[olive,fill=olive!30!white]
        coordinates{(HotpotQA,0.434) (ANLI,0.586) (SQuAD,0.581) (OpenPI,0.364)};
        \addplot[red,fill=red!30!white]
        coordinates{(HotpotQA,0.449) (ANLI,0.551) (SQuAD,0.579) (OpenPI,0.364)};
        \addplot[orange,fill=orange!30!white]
        coordinates{(HotpotQA,0.460) (ANLI,0.563) (SQuAD,0.594) (OpenPI,0.358)};    
    \end{axis}
\end{tikzpicture}
\begin{tikzpicture}
    \begin{axis}[ybar,width=3.8cm,height=3.25cm,enlarge x limits=0.5,symbolic x coords={CNN/DM,XSUM},xtick=data,xticklabel style={text width=1.5cm,font=\tiny,align=center,max space between ticks=12,major tick length=0.125cm,},yticklabel style={overlay,font=\tiny},legend style={font=\tiny},bar width=5pt,]
        \addplot[purple,fill=purple!30!white]
        coordinates{(CNN/DM,10.9) (XSUM,11.2)};
        \addplot[olive,fill=olive!30!white]
        coordinates{(CNN/DM,11.8) (XSUM,11.0)};
        \addplot[red,fill=red!30!white]
        coordinates{(CNN/DM,11.7) (XSUM,11.0)};
        \addplot[orange,fill=orange!30!white]
        coordinates{(CNN/DM,12.3) (XSUM,11.7)};
   \end{axis}
\end{tikzpicture}
\begin{tikzpicture}
   \begin{axis}[ybar,width=8.75cm,height=3.25cm,enlarge x limits=0.1,legend style={at={(0.5,-0.5)},anchor=north,legend columns=4},symbolic x coords={HellaSWAG,wikiHow goal-step,wikiHow temporal,WinoGrande,Yelp,IMDb},xtick=data,xticklabel style={text width=1.5cm,font=\tiny,align=center,max space between ticks=8,major tick length=0.1cm,ytick placement tolerance=-0.2},yticklabel style={overlay,font=\tiny,},legend style={font=\tiny},bar width=5pt,]
        \addplot[purple,fill=purple!30!white]
        coordinates{(HellaSWAG,0.495) (wikiHow goal-step,0.709) (wikiHow temporal,0.630) (WinoGrande,0.455) (Yelp,0.900) (IMDb,0.951)};
        \addplot[olive,fill=olive!30!white]
        coordinates{(HellaSWAG,0.571) (wikiHow goal-step,0.895) (wikiHow temporal,0.690) (WinoGrande,0.719) (Yelp,0.901) (IMDb,0.949)};
        \addplot[red,fill=red!30!white]
        coordinates{(HellaSWAG,0.606) (wikiHow goal-step,0.898) (wikiHow temporal,0.727) (WinoGrande,0.716) (Yelp,0.907) (IMDb,0.951)};
        \addplot[orange,fill=orange!30!white]
        coordinates{(HellaSWAG,0.487) (wikiHow goal-step,0.787) (wikiHow temporal,0.768) (WinoGrande,0.684) (Yelp,0.900) (IMDb,0.941)};
        \legend{\texttt{Vanilla}, \texttt{VI}, \texttt{VIC}, \texttt{CVIC}}
    \end{axis}
\end{tikzpicture}
\caption{Comparison of \texttt{code-davinci-002} across the four types of code prompts. Figures are split to allow for different y-axis scales. We see that different prompts do better on different tasks and while some tasks have high variance over prompt types, others do not.}
\label{fig:code_prompts}
\end{figure}

\minisection{What is the best type of code prompt?}
We compare performance across the four code prompt types from Section~\ref{sec:experimental_design} on all 12 tasks using \texttt{code-davinci-002} and report our results in Figure~\ref{fig:code_prompts}. We find that no single type of code prompt performs significantly better than the others across all tasks and that the relative difference in performance between code prompts also varies significantly across tasks. For example, on IMDb and SQuAD all code prompts have roughly even performance while for tasks such as wikiHow-Temporal and WinoGrande we see a near 14\% accuracy difference between the worst and best prompt. 

In Appendix~\ref{sec:rank-based-stats}, we calculate the average rank of each code prompt type relative to each other and find that the ``Var Identifier + Comments'' (\texttt{VIC}) prompt is the best across all tasks on average (2.25 avg. rank). We thus use this prompt type for our comparison in all future sections.

\begin{figure}
    \centering
    \includegraphics[width=\columnwidth]{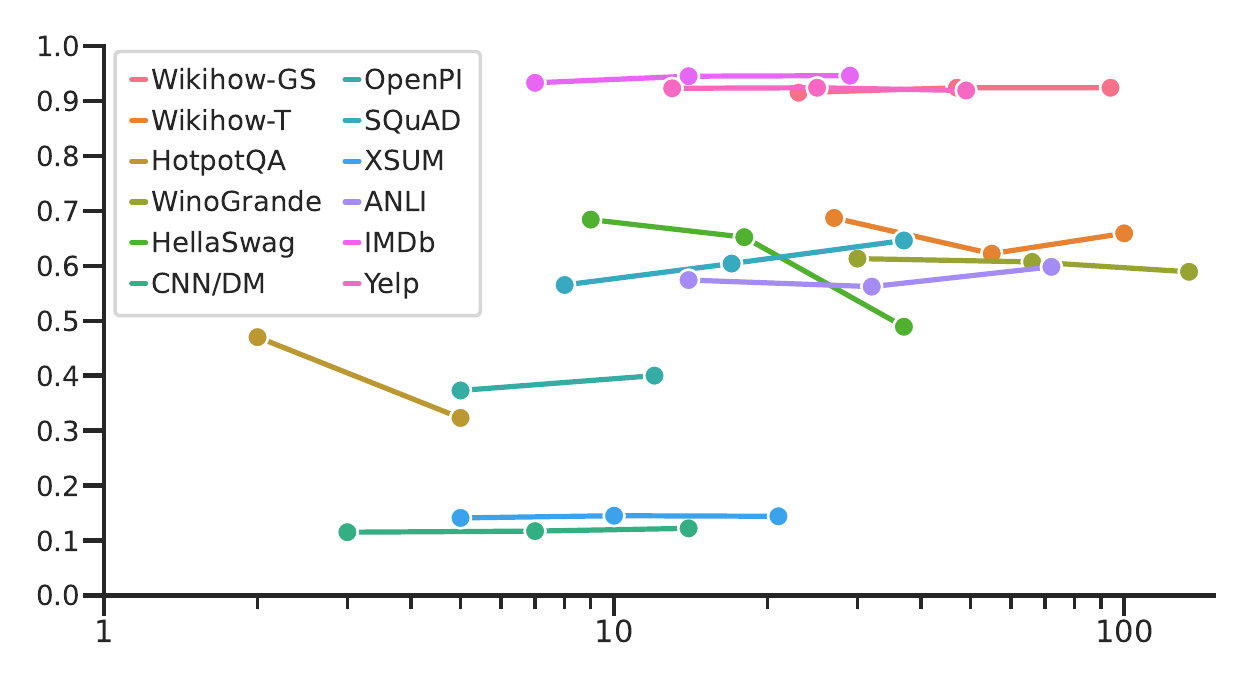}
    \caption{Performance score (y-axis) vs number of in-context examples (x-axis, in log scale) using code prompts (\texttt{VIC}) with \texttt{code-davinci-002}. We see that increasing number of examples does not always increase performance and in some cases makes it worse.}
    \label{fig:prompt_lengths}
\end{figure}

\begin{table*}
    \small
    \centering
    \begin{tabular}{l|l|c|c|c|c|c|c|c|c|c}
    \toprule
        Dataset & Metric & \multicolumn{3}{c|}{\texttt{davinci}} & \multicolumn{3}{c|}{\texttt{code-002}} & \multicolumn{3}{c}{\texttt{text-002}}\\
        & & +Text & +Code & $\Delta$ & +Text & +Code & $\Delta$ & +Text & +Code &$\Delta$\\
        \midrule
        Hellaswag &Accuracy&0.321&0.307&{\color{lightred}-0.014}&0.652&0.606&{\color{lightred}-0.046}&0.717&0.773&{\color{cyan}+0.046}\\
        wikiHow goal-step&Accuracy&0.347&0.302&{\color{lightred}-0.045}&0.924&0.898&{\color{lightred}-0.026}&0.919&0.915&{\color{lightred}-0.004}\\
        wikiHow temporal&Accuracy&0.495&0.532&{\color{cyan}+0.037}&0.622&0.727&{\color{blue}+0.105}&0.688&0.761&{\color{cyan}+0.073}\\
        Yelp&Pearson $\rho$&0.913&0.896&{\color{lightred}-0.017}&0.924&0.907&{\color{lightred}-0.017}&0.919&0.904&{\color{lightred}-0.015}\\
        IMDb&Accuracy&0.872&0.935&{\color{cyan}+0.063}&0.945&0.951&{\color{cyan}+0.006}&0.940&0.952&{\color{cyan}+0.012}\\
        WinoGrande&Accuracy&0.513&0.500&{\color{lightred}-0.013}&0.607&0.716&{\color{blue}+0.109}&0.628&0.726&{\color{cyan}+0.098}\\
        ANLI&Accuracy&0.333&0.360&{\color{cyan}+0.027}&0.562&0.551&{\color{lightred}-0.011}&0.504&0.557&{\color{cyan}+0.053}\\
        HotpotQA&Macro-F1&-&-&-&0.470&0.449&{\color{lightred}-0.021}&0.490&0.350&{\color{lightred}-0.140}\\
        SQuAD&Macro-F1&0.482&0.466&{\color{lightred}-0.016}&0.604&0.579&{\color{lightred}-0.025}&0.670&0.656&{\color{lightred}-0.014}\\
        OpenPI&ROUGE-F1&-&-&-&37.33&36.36&{\color{lightred}-0.970}&35.60&31.30&{\color{darkred}-4.300}\\
        CNN/Daily Mail&ROUGE-2&9.28&9.13&{\color{lightred}-0.150}&11.74&11.67&{\color{lightred}-0.070}&13.63&13.55&{\color{lightred}-0.080}\\
        XSUM&ROUGE-2&9.38&6.83&{\color{darkred}-2.550}&14.51 &11.03&{\color{darkred}-3.580}&14.48&13.26&{\color{darkred}-1.220}\\
        \bottomrule
    \end{tabular}
    \caption{Performance of the three LMs when using code prompts (+Code) vs. using text prompts (+Text). Blank cells indicate tasks for which single test examples could not fit in the context window. Color indicates whether or not code prompts are {\color{blue}better}, {\color{cyan}slightly better}, {\color{lightred}slightly worse}, or {\color{darkred}worse} than text prompts. We see that while code prompts outperform text prompts for certain tasks (such as wikiHow temporal and WinoGrande) text prompts are better on average. We also find that instruction fine-tuning (\texttt{text-002}) allows for better code prompt utilization.}
    \label{tab:textvscode}
\end{table*}

\paragraph{How many in-context examples should we include in our code prompt?}
We would like to also investigate how the number of in-context examples in the prompt affects models' ability to perform the task. We therefore conducted an experiment where we filled the context window of \texttt{code-davinci-002} with in-context examples up to 2000 tokens, 4000 tokens, and 8000 tokens and plotted the validation accuracy of the model with respect to the number of examples in Figure~\ref{fig:prompt_lengths}. 

Contrary to expectations, we find that the number of in-context examples has little effect on model performance for most tasks and actually has a \textit{negative} effect on some tasks. This is especially interesting given that previous work on in-context learning with text prompts finds roughly monotonic improvement from adding more in-context examples \cite{liu2021makes}. While further research is necessary, it seems that code prompts may have different scaling behavior than text prompts when used in in-context learning. 

\paragraph{Which is better: code or text prompts?}
In our main experiment we compare the performance of the three GPT models on code prompts (\texttt{VIC} style) and text prompts across the 12 datasets. Given the results from Figure~\ref{fig:prompt_lengths}, we fill the context window of all models with in-context examples up to 4000 tokens to serve as a middle ground for comparing code and text prompts. We report the results of our main experiment in Table~\ref{tab:textvscode} and see several surprising trends.

First, we find that prompting PLMs with code leads to substantial increases in performance for certain few reasoning tasks but that this trend does not hold across all tasks---or even all reasoning tasks. For example, when using code prompts with \texttt{code-davinci-002}, we see a 10.5\% accuracy increase on wikiHow temporal ordering but a 2.6\% accuracy decrease on wikiHow goal-step inference despite both being commonsense reasoning tasks and having identical source material.

Second, we find that supervised instruction fine-tuning on natural language demonstrations does not hurt model performance on code. Rather, we instead observe that code prompts outperform text prompts on \textit{more} tasks when using \texttt{text-davinci-002} than when using \texttt{code-davinci-002} despite the fact that \texttt{text-davinci-002} received no additional fine-tuning on code instructions.

Finally, we find that LMs not explicitly trained on code can also benefit from code prompting on certain reasoning tasks. In particular, code prompts outperform text prompts on \texttt{davinci} for 3 out of our 12 tasks---the same proportion as \texttt{code-davinci-002}. The tasks that benefit from code prompts also seem to be largely consistent across the three types of models tested, suggesting some underlying trend as to which tasks systematically benefit from structured input.

\section{Conclusion}
In this work we investigate whether or not there exists a systematic performance difference between prompting PLMs with code or with text. We confirm that there are indeed tasks for which code prompting is significantly more effective than text prompting and that this finding holds across different types of models. However, for most tasks, we find that text prompting is still the best method for eliciting few-shot generalization from PLMs. 

Given this result it seems reasonable to attempt to predict which tasks will benefit from code prompts and which tasks will not. However, we show that making such predictions based on simple heuristics such as domain and task category is difficult and that the larger trends remain unclear. Future work should seek to investigate the core mechanism behind what makes code prompting effective for certain tasks.

Finally, concurrent to our work, a new line of research has emerged wherein models generate code and \textit{execute} that code to produce valid output \cite{chen2022program,mishra2022lila,gao2022pal,lyu-etal-2023}. Future work should considering whether or not the tasks that benefit from executable code prompts and non-executable code prompts have any overlap. 

\section*{Limitations}
One significant limitation to our study is that, as of March 23rd 2023, OpenAI has deprecated access to \texttt{code-davinci-002}\footnote{\url{https://platform.openai.com/docs/model-index-for-researchers}}, thus rendering our results non-replicable for any team not granted special access to these models by OpenAI. We did not anticipate this deprecation while conducting this work and we believe this raises serious questions about the usage of API-based language models in scholarly work.

Another limitation is that the 12 tasks we selected may not be representative of the broader population of natural language tasks. Had we conducted our experiments on a larger selection of tasks there may have been larger-scale trends that we would have been able to uncover.

The largest and most pressing limitation with our work is that the models we are testing on have closed-source pre-training datasets. Thus, we are unable to verify the extent to which our task datasets have been included in the training or instruction fine-tuning data. Given that the training data for most of the models tested in this work cuts off in late 2021, this is a very strong possibility. Our results should be viewed with this limitation strongly in mind.

Finally, while we experimented with different code prompts, the search space of possible prompts is very large. Thus, it is very likely that there exists some prompt that outperforms our chosen prompts for each task. Drawing conclusions based on a limited sampling of prompts is tenuous and while methods exist for searching the space of all prompts, such techniques lack interpretability and erase any distinction between code and text prompt \cite{li2021prefixtuning}.

\section*{Acknowledgements}
The paper is dedicated to the late Prof. Dragomir Radev, the first mentor in NLP of the author Li Zhang, for igniting his passion for research and passing onto him much knowledge.

We thank Shuyan Zhou, Aman Madaan, and Niket Tandon for valuable discussions about this work and we thank Alyssa Hwang for her contributions to the structure, presentation, and narrative of the final paper.

This research is based upon work supported in part by the DARPA KAIROS Program (contract FA8750-19-2-1004), the DARPA LwLL Program (contract FA8750-19-2-0201), the Office of the Director of National Intelligence (ODNI) via the IARPA HIATUS Program (contract 2022-22072200005), the NSF (Award 1928631), and gifts from Roblox and Salesforce. Approved for Public Release, Distribution Unlimited. The views and conclusions contained herein are those of the authors and should not be interpreted as necessarily representing the official policies, either expressed or implied, of DARPA, ODNI, IARPA, NSF, the U.S. Government, or of Roblox or Salesforce. The U.S. Government is authorized to reproduce and distribute reprints for governmental purposes notwithstanding any copyright annotation therein.

\bibliography{paper}
\bibliographystyle{acl_natbib}

\appendix

\section{Detailed Task Description}
\label{sec:detailed-tasks}
\paragraph{Summarization} is the task of composing a concise description of a lengthy text. Given a long narrative, the model is tasked with composing a short summary that contains the salient events in the original text. 

For our study, we select the \textit{CNN/Daily Mail} \cite{hermann2015teaching, nallapati2016abstractive} and \textit{XSUM} \cite{narayan2018don} datasets as both are variants on the challenging abstractive summarization task. \textit{XSUM} tasks models with generating extremely concise 1 to 2 sentence summaries of news articles and \textit{CNN/Daily Mail} tasks models with generating reasonably concise but longer abstractive summaries. For both \textit{CNN/Daily Mail} and \textit{XSUM} datasets, we use ROUGE-2 score for evaluation.

\paragraph{Question Answering (QA)} is the task of composing answers given a question and an optional context passage. When this context passage is provided the task is referred to as ``open-book'' QA and when it is not it is referred to as ``closed-book'' QA. Open-book QA tasks examine language models' ability to understand and extract information from their context while Closed-book QA tasks evaluate the amount of knowledge encapsulated in language models during pre-training.

For our study we pick two open-book QA datasets, \textit{SQuADv2} \cite{rajpurkar2018know} and \textit{HotpotQA} \cite{yang2018hotpotqa}, which allow us to focus our evaluation on how structured prompts affect models' ability to comprehend long text input.

For both \textit{SQuADv2} and \textit{HotpotQA}, we evaluate model performance based on the macro-averaged F1 score as proposed in \citet{rajpurkar2016squad}. This metric measures the average overlap between the prediction and ground truth answer. It is calculated by treating the prediction and ground truth as bags of tokens, and first computing their F1. Then, the maximum F1 score is taken over all of the ground truth answers for a given question, and that score is averaged over all of the questions to get the final result.

\paragraph{Commonsense Reasoning} is a machine reasoning task that demands the use of commonsense knowledge which is oftentimes implicitly present in the text \cite{sap2020introductory}. The customary formulation of commonsense reasoning tasks are \textit{Classification}, where the input is a context, optionally with candidate answers as choices, and the output is a label from a pre-defined label space, and \textit{Question Answering} (QA), where the input is a context followed by a reasoning question and the output is in free-form language.   

In this study, we selected four Classification style commonsense reasoning tasks: \textit{wikiHow Temporal} and \textit{wikiHow Goal-Step} \cite{zhang2020reasoning}, \textit{ANLI} \cite{nie2020adversarial}, and \textit{HellaSwag} \cite{zellers2019hellaswag}. We also included one Question Answering style task with \textit{OpenPI} \cite{tandon2020dataset}. In addition, we evaluate our models on \textit{WinoGrande} a comprehensive reasoning benchmark dataset \cite{sakaguchi2021winogrande}. 

For \textit{wikiHow Goal-Step}, \textit{wikiHow Temporal}, \textit{HellaSwag}, \textit{WinoGrande}, and \textit{ANLI}, we use classification accuracy as the evaluation metric. To evaluate \textit{OpenPI}, we use F1 score based on the ROUGE metric as described in the original paper \cite{tandon2020dataset}.

\paragraph{Sentiment Analysis} is a task that is concerned with judging emotion and its degree in text. Given a passage, a language model is tasked with classifying the sentiment (positive, negative, neutral) and/or its degree (strongly, weakly, moderately). 

The selected datasets, namely \textit{IMDb} \cite{maas2011learning} and \textit{Yelp} \cite{zhang2015character}, are both constructed using customer reviews. The IMDb dataset proposes a binary classification problem where the input is a movie review and the label space is $\{negative, positive\}$. Yelp proposes a five-way classification problem where the input is a restaurant review and the label space is the number of stars (out of 5) the customers assigned to the restaurant. 

For \textit{IMDb}, we use accuracy as the evaluation metric and for \textit{Yelp}, we use Pearson Correlation between the predicted rating and the ground truth rating as the evaluation metric.

\begin{table}
    \small
    \centering
    \begin{tabular}{c|c|c|c|c}
    \toprule
    &\texttt{Vanilla} & \texttt{VI} & \texttt{VIC} & \texttt{CVIC} \\
    \midrule
    HellaSwag&3&2&1&4 \\
    wikiHow Goal-Step&4&2&1&3 \\
    wikiHow Temporal&4&3&2&1 \\
    Yelp&4&2&1&4 \\
    IMDb&1&3&1&4 \\
    WinoGrande&4&1&2&3 \\
    HotpotQA&4&3&2&1 \\
    ANLI&1&2&4&3 \\ 
    OpenPI&1&2&3&4 \\ 
    SQuAD&1&3&4&2 \\
    CNN/Daily Mail&4&2&3&1 \\
    XSUM&2&4&3&1 \\
    \midrule
    \midrule
    \textit{Mean}& 2.75 & 2.42 & 2.25 & 2.58 \\
    \textit{Standard Deviation}& 1.36 & 0.76 & 1.09 & 1.26 \\
    \bottomrule
    \end{tabular}
    \caption{Relative performance rank of the four code prompt types from Section~\ref{sec:experimental_design} across the 12 tasks. Ranks are calculated based on the results reported in Figure~\ref{fig:code_prompts}. We see that the ``Variable Identifier + Comments'' (\texttt{VIC}) style prompt performs the best out of all code prompt types on average.}
    \label{tab:rank-based-stat}
\end{table}

\section{Ranking of Code Prompt Styles}
\label{sec:rank-based-stats}
In Table~\ref{tab:rank-based-stat} we report the rank-based statistics of the four code prompt types from Section~\ref{sec:experimental_design} on our 12 tasks. Ranks are calculated based on the results reported in Figure~\ref{fig:code_prompts} of the main paper. The numbers in a row reflect the relative standing of each code prompt on the corresponding task. While we note that all code prompts perform within $\pm$0.5 ranks of each other on average, we see that on average the \texttt{VIC} prompt performs the best across all tasks and the \texttt{Vanilla} prompt performs the worst. Looking to the standard deviation section, we see that the \texttt{VI} prompt performs the most consistently across all tasks and that once again the \texttt{Vanilla} prompt performs the least consistently.

\section{Ablation Study}
\label{app:ablation_study}
To see whether the findings in our Results section could be attributed to variance in the random sampling of in-context training examples per test example, we conduct five repeated runs using \texttt{code-davinci-002} with different random seeds each time and calculated the standard deviation across the five runs. We report our results in Table~\ref{tab:repeated-run-stats} and find that the choice of in-context examples accounts for very little of the observed variance across prompt type and context length. This finding is surprising as previous work has shown that the selection and ordering of in-context examples has a very large effect on the performance of models \cite{liu2021makes}. However, it seems that our approach of random sampling in-context examples per test item helps to lessen this inherent variance.

\begin{table}
    \small
    \centering
    \begin{tabular}{l|c|c}
    \toprule
        Dataset &Performance&$\sigma$\\
        \midrule
        Hellaswag&0.65, 0.67, 0.69, 0.67, 0.67&$\pm$0.01\\ 
        wikiHow-GS&0.51, 0.51, 0.51, 0.50, 0.51&$\pm$0.00\\ 
        wikiHow-T&0.62, 0.65, 0.63, 0.63, 0.62&$\pm$0.01\\ 
        Yelp&0.92, 0.92, 0.92, 0.92, 0.92&$\pm$0.00\\ 
        IMDb&0.94, 0.94, 0.94, 0.94, 0.94&$\pm$0.00\\ 
        WinoGrande&0.62, 0.64, 0.61, 0.62, 0.62&$\pm$0.01\\ 
        HotpotQA&0.35, 0.33, 0.35, 0.35, 0.35&$\pm$0.01\\ 
        ANLI&0.59, 0.58, 0.57, 0.60, 0.61&$\pm$0.01\\ 
        OpenPI&36.3, 38.1, 38.3, 37.7, 39.9&$\pm$1.16\\ 
        SQuAD&0.60, 0.62, 0.61, 0.60, 0.63&$\pm$0.01\\ 
        CNN/DM&11.7, 12.0, 12.4, 12.3, 12.0&$\pm$0.25\\ 
        XSUM&14.5, 14.9, 15.5, 15.2, 15.4&$\pm$0.36\\ 
        \bottomrule
    \end{tabular}
    \caption{Comparison across 5 repeated runs of the \texttt{code-davinci-002} model with text prompts using different random seeds for sampling in-context examples. We see minimal standard deviation ($\sigma$) between the runs.}
    \label{tab:repeated-run-stats}
\end{table}

\section{Evaluation on \texttt{text-davinci-003}}
\label{app:text_003}
While conducting our research into the differences between code and text prompts, OpenAI released the \texttt{text-davinci-003} model. This model differs from \texttt{text-davinci-002} in that it is trained using Reinforcement Learning with Human Feedback (RLHF) instead of supervised instruction fine-tuning \cite{ouyang2022training}. Out of curiosity, to see the effect of this new training paradigm, we conducted experiments comparing this new \texttt{text-davinci-003} model to the other GPT-3.5 models (\texttt{text-davinci-002} and \texttt{code-davinci-002}). We report the results of our comparison across the 12 evaluation tasks in Table~\ref{tab:003-comparison}. 

We see that while \texttt{text-davinci-003} out-performs all previous models on \textit{wikiHow Temporal}, \textit{WinoGrande}, and \textit{OpenPI}, it does significantly worse than previous models on \textit{wikiHow Goal-Step} and \textit{HotpotQA}. Such large reductions in performance are to be somewhat expected when using RLHF given the costly nature of collecting human demonstrations. However, the magnitude of the decreases (-50.1\% for \textit{wikiHow} and -11.2\% for \textit{HotpotQA}) is nonetheless surprising and such results raise important questions about exactly what is being learned when conducting instruction fine-tuning and whether or not this learned information can generalize to tasks not seen during fine-tuning.

\begin{table}
     \small
     \centering
     \begin{tabular}{l|c|c|c}
     \toprule
     Task&\texttt{code-002}&\texttt{text-002}&\texttt{text-003}\\
     & (base) & (+IFT) & (+RLHF)\\
     \midrule
     HellaSwag&0.652&\textbf{0.717}&0.714\\
     wikiHow GS&\textbf{0.924}&0.919&0.510\\
     wikiHow T&0.622&0.688&\textbf{0.815}\\
     Yelp&\textbf{0.924}&0.919&0.903\\
     IMDb&\textbf{0.945}&0.940&0.938\\
     WinoGrande&0.607&0.628&\textbf{0.735}\\
     ANLI&\textbf{0.562}&0.504&0.549\\
     HotpotQA&0.470&\textbf{0.490}&0.378\\
     SQuAD&0.604&\textbf{0.670}&0.663\\
     OpenPI&37.33&35.60&\textbf{39.06}\\
     CNN/DM&11.74&\textbf{13.63}&12.64\\
     XSUM&\textbf{14.51}&14.48&13.36\\  
     \bottomrule
     \end{tabular}
     \caption{Performance of the three GPT-3.5 models across our 12 datasets with \textbf{text prompts}. (+IFT) indicates the addition of supervised instruction fine-tuning and (+RLHF) indicates the addition of training using Reinforcement Learning from Human Feedback \cite{ouyang2022training}. We see that RLHF does not always improve performance and that for some tasks (HotpotQA and wikiHow Goal-Step) it causes large degradations in performance.}
     \label{tab:003-comparison}
\end{table}

\section{Evaluation Cost}
\label{app:eval_cost}
In this section we report the approximate cost of conducting our experiments. In our study we use four OpenAI models, namely \texttt{davinci}, \texttt{code-davinci-002}, \texttt{text-davinci-002} and \texttt{text-davinci-003}. While \texttt{code-davinci-002} is free to use at the time of this study, we report the approximate cost of running the experiments on the other three models\footnote{The cost of querying \texttt{davinci}, \texttt{text-davinci-002} and \texttt{text-davinci-003} is \$0.02/1,000 tokens at the time of study. See \url{https://openai.com/pricing} for more details.} in Table~\ref{tab:est_cost}. To estimate the cost of an experiment, we calculate the approximate number of tokens necessary for computing one dataset example and then multiplied that by the number of examples in the dataset. For classification tasks, since we fill up the context window to roughly 4000 tokens for every test example, we estimate the number of tokens to be 4000 (3999 tokens for the prompt and 1 token for the label). To estimate cost for generative tasks (OpenPI, HotpotQA, SQuAD, CNN/Daily Mail, and XSUM), we compute the average generation length from our generated samples and assume the in-context examples take up 3500 tokens. While this calculation results in a fairly loose upper bound, we believe this to be a good estimate of the total cost incurred by the project as such overestimates help offset the cost of other miscellaneous API queries done over the course of the project.

\begin{table}
    \small
    \centering
    \begin{tabular}{l|l|l}
    \toprule
    Dataset & Num. Examples & Est. Cost \\
    \midrule
    HellaSwag &1000 / 10042&\$240.48\\
    wikiHow Goal-Step&1000 / 1073&\$240.48\\
    wikiHow Temporal &1000 / 3100&\$240.48\\
    WinoGrande &1000 / 1767&\$240.48\\
    OpenPI &111 / 111&\$28.08\\
    ANLI  &1000 / 3000&\$240.48\\
    Yelp &1000 / 10000&\$240.48\\
    IMDb &1000 / 25000 &\$240.48\\
    HotpotQA &1000 / 7405 &\$241.20\\
    SQuAD &1000 / 11873 &\$241.08\\
    CNN/Daily Mail &1000 / 13368&\$257.91\\
    XSUM &1000 / 11332&\$246.66\\
    \midrule 
    \midrule
    \multicolumn{2}{c|}{Total Cost} & \$2698.29\\
    \bottomrule
    \end{tabular}
    \caption{The total estimated cost of running \texttt{davinci}, \texttt{text-davinci-002} and \texttt{text-davinci-003} for 1000 data samples from each dataset (except for OpenPI).}
    \label{tab:est_cost}
\end{table}

\end{document}